\documentclass{article}
\usepackage{ijcai21}

\usepackage{times}
\usepackage{soul}
\usepackage{url}
\usepackage[hidelinks]{hyperref}
\usepackage[utf8]{inputenc}
\usepackage[small]{caption}
\usepackage{graphicx}
\usepackage{amsmath}
\usepackage{amsthm}
\usepackage{booktabs}
\usepackage{algorithm}
\usepackage{algorithmic}
\usepackage{todonotes}
\usepackage{xspace}
\usepackage{booktabs}
\usepackage{makecell}
\usepackage{array,multirow,graphicx}
\newcommand{\code}[1]{\textbf{#1}}

\newcommand{\rel}[3]{$\langle #1,\code{#2},#3 \rangle$}

\newcommand{\hide}[1]{}

\newcommand{\spear}{SpEAR\xspace}
\newcommand{\spert}{SpERT\xspace}

\newcommand{\secref}[1]{Section~\ref{sec:#1}}

\newcommand{\figref}[1]{Figure~\ref{fig:#1}}

\newcommand{\tabref}[1]{Table~\ref{tab:#1}}

\newcommand{\textq}[1]{``#1''}

\title{Extracting Qualitative Causal Structure with Transformer-Based NLP}

\author{
Scott E. Friedman$^1$\footnote{Contact Author}\And%
Ian H. Magnusson$^{1,2}$\And%
Sonja M. Schmer-Galunder$^1$
\affiliations
$^1$SIFT, Minneapolis, MN, USA\\
$^2$Northeastern University, MA, USA\\
\emails
\{sfriedman, imagnusson, sgalunder\}@sift.net
}

\begin{document}

\maketitle

\begin{abstract}

Qualitative causal relationships compactly express the direction, dependency, temporal constraints, and monotonicity constraints of discrete or continuous interactions in the world.
In everyday or academic language, we may express interactions between quantities (e.g., sleep decreases stress), between discrete events or entities (e.g., a protein inhibits another protein's transcription), or between intentional or functional factors (e.g., hospital patients pray to relieve their pain).
This paper presents a transformer-based NLP architecture that jointly identifies and extracts (1) variables or factors described in language, (2) qualitative causal relationships over these variables, and (3) qualifiers and magnitudes that constrain these causal relationships.
We demonstrate this approach and include promising results from in two use cases, processing textual inputs from academic publications, news articles, and social media.

\end{abstract}


\section{Introduction}
\label{sec:intro}

We express qualitative causal relationships in our everyday language and our scientific texts to capture the relationship between quantities or entities or events, compactly communicating how one event or purpose or quantity might be affected by another.
These causal relations are not complete mechanism descriptions in themselves, but we use them frequently in everyday language and formal instruction to express causality, allowing us to avoid unnecessary detail or to hedge when details are uncertain.

Identifying these causal relationships from natural language---and also properly identifying the factors that they relate---remains a challenge for NLP systems.
This difficulty is due in part to the expressiveness of our language, e.g., the multitude of ways we may describe how an experimental group scored higher on an outcome than a control group, and also due to the complexity of the systems we describe.

This paper describes an approach to automatically extracting (1) entities that are the subject of causal relationships, (2) qualitative causal relationships describing mechanisms, purposes, monotonicity, and temporal priority, and (3) multi-label attributes to further characterize the causal structure.
Our primary claim is that context-sensitive language models can detect and characterize the qualitative causal structure of everyday and scientific language.
As evidence, we present our \spear transformer-based NLP model based on BERT \cite{devlin-etal-2019-bert} and \spert \cite{eberts2019span} that extracts causal structure from text as knowledge graphs, and we present promising initial results on (1) characterizing scientific claims and (2) representing and traversing descriptive mental models from ethnographic texts.

The causal, semantic graphs produced by \spear do not conform to a strict ontology and therefore do not presently support the formal reasoning afforded by ontologies with strict selectional restrictions and relational constraints; however, we demonstrate that these outputs allow traversal across concepts to characterize meaningful causal influences.

We continue with a review of related work in qualitative causal representations (\secref{qr}) and transformer-based NLP (\secref{nlp}).
We then describe our approach (\secref{approach}) and preliminary results in two domains (\secref{results}).
We conclude with a discussion of future work in this area.

\begin{figure*}[htb]
\centering
\includegraphics[width=.9\linewidth]{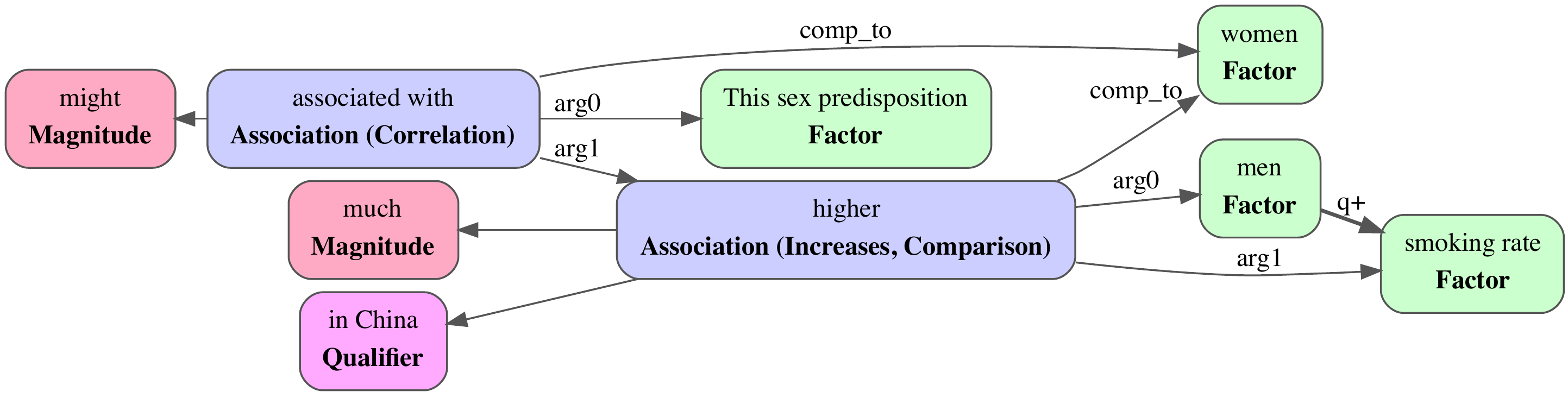}
\vspace{-0.1in}
\caption{\spear knowledge graph output for the text \textq{This sex predisposition might be associated with the much higher smoking rate in men than in women in China.} Includes a correlation, a comparison with a qualitative increase, magnitudes, and a location qualifier.}
\label{fig:score1}
\vspace{0.3in}
\includegraphics[width=.9\linewidth]{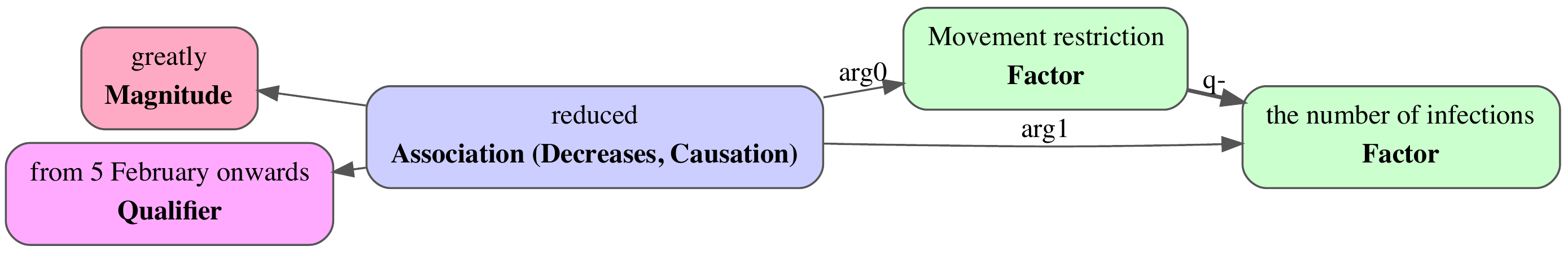}
\caption{\spear knowledge graph output for \textq{Movement restriction greatly reduced the number of infections from 5 February onwards.} Includes a causal association, a qualitative decrease, a magnitude, and a temporal qualifier.}
\label{fig:score2}
\end{figure*}

\section{Background and Related Work}

We review related work in representing causal relations, which informs the present approach.
We then review previous work in transformer-based NLP, including the \spert system \cite{eberts2019span} which is a subsystem of our architecture.

\subsection{Qualitative Causal Relations}
\label{sec:qr}

The knowledge representations described in this paper are motivated by previous work in qualitative reasoning and simulation \cite{forbus2019qualitative}.
For example, \emph{qualitative proportionalities} describe how one quantity impacts another, in a directional, monotonic fashion.
In this work, we designate \rel{a}{q+}{b} (and respectively, \rel{a}{q–}{b}) as qualitative proportionalities from $a$ to $b$, such that increasing $a$ would increase (and respectively, decrease) $b$.
This is motivated by formal quantity-to-quantity $\alpha_{Q+/-}$ relations \cite{forbus1984qualitative} and $M^{+/-}$ relations in qualitative simulation \cite{kuipers1986qualitative}, but our semantics are less constrained than either of these, due to tendencies in language to express an increase from an event to a quantity (e.g., \textq{smoking that cigarette will increase your risk of cancer}) or from entities to activities (e.g., \textq{the prime increased participants' retrieval of the cue}), and so on.

Previous work in philosophy \cite{dennett1989intentional} and cognitive psychology \cite{lombrozo2006functional} has acknowledged intentional (i.e., psychological, goal-based) and teleological (i.e., functional, design-based) relationships as types of causal relations.
Previous work has represented these as lexical qualia or affordances \cite{pustejovsky1991syntax}.
In this work, we represent purposeful, intentional actions as a qualitative relationship \rel{a}{forPurpose}{b}, such that the actor of action $a$ may have intended the purpose or goal $b$, e.g., \textq{they prayed for a safe pregnancy.}
We represent teleological (i.e., functional or design-based) causal relations as \rel{a}{hasFunction}{b} to indicate that the action or artifact $a$ is designed or otherwise has a function to achieve $b$, e.g., \textq{the artifacts provide protection for pregnant women.}

\subsection{Causal and Transformer-Based NLP}
\label{sec:nlp}

Transformer-based methods for NLP utilize neural networks to encode a sequence of textual tokens (i.e., words or sub-words) into large vector-based representations for each token, sensitive to the context of the surrounding tokens \cite{devlin-etal-2019-bert}.
This is widely regarded as a state-of-the-art methodology for NLP, and has been used to process text to extract knowledge graphs, e.g., of people and relations \cite{eberts2019span}.
The architecture we present in this paper has been applied to datasets of scientific claims \cite{magnusson2021score} and hate speech \cite{tybalt2021cogsci}.
Many existing transformer models---similar to the architecture presented in this paper---require hundreds (sometimes thousands) of labeled training examples to reach high proficiency.

Previous symbolic semantic parsers extract scientific claims and assertions from text with explicit relational knowledge representations \cite{allen2015complex}, but many of these approaches rely on rule-based engines with hand tuning, which requires NLP experts to maintain and adapt to new domains.
By contrast, our approach is designed to extract causal relationships at a level of expressiveness comparable to these symbolic systems while using the advances in transformer-based models such as \spert \cite{eberts2019span} to learn graph-based representation from examples alone.

Other NLP approaches use machine learning to extract features from scientific texts, e.g., to identify factors and directions of influence \cite{mueller2019deepcause}; however these approaches do not explicitly infer the relations between elements in the claim, as shown with the present approach.

\section{Approach}
\label{sec:approach}

We describe our graph schema for representing the entities, attributes, and qualitative relationships extracted from text.  
We discuss the general problem definition and then we explain the specific graph schemas in two domains: (1) scientific claims and (2) ethnographic mental models.  

\subsection{Knowledge Graphs}

The \spear knowledge graph format include the following three types of elements: entities, attributes, and relations.
We describe each of these before defining the problem and describing the architecture.

\begin{figure*}[htb]
\centering
\includegraphics[width=.7\linewidth]{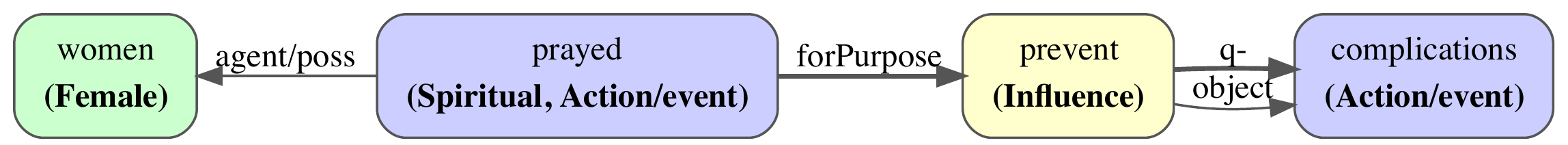}
\vspace{-0.1in}
\caption{\spear knowledge graph for \textq{Therefore, the women prayed to prevent any complications,} including \code{forPurpose} and \code{q-} relations.}
\label{fig:hab1}
\vspace{0.3in}
\includegraphics[width=.65\linewidth]{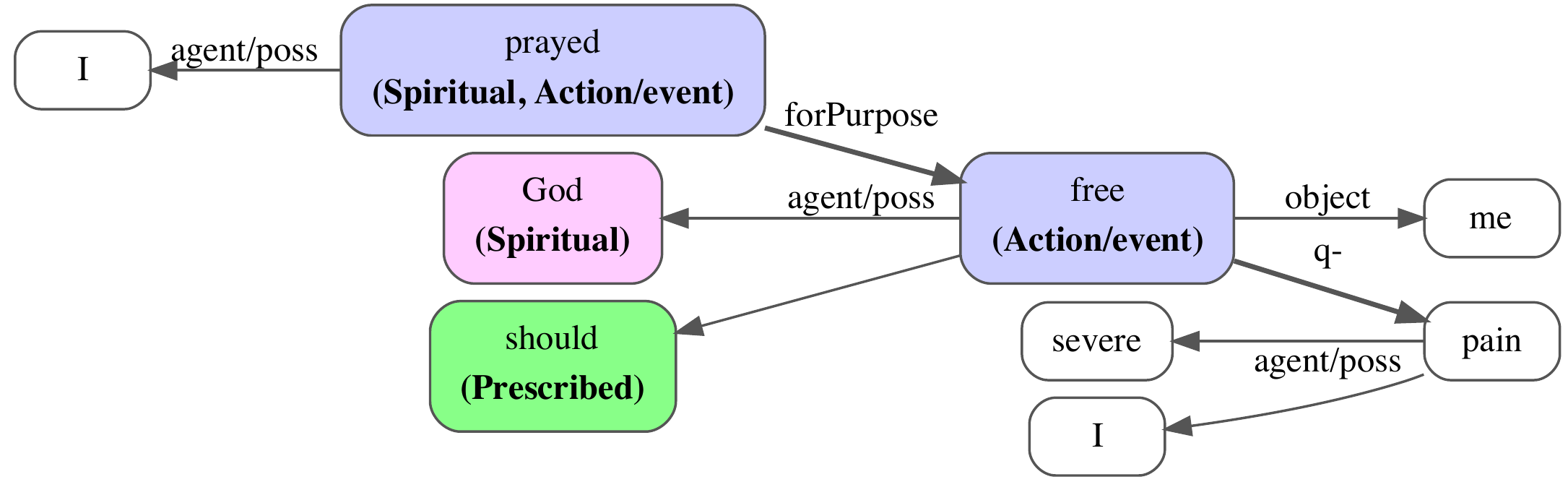}
\caption{\spear knowledge graph for \textq{I prayed that God should free me from the severe pain I was going through.}}
\label{fig:hab2}
\end{figure*}

\begin{figure}[htb]
\centering
\includegraphics[width=\linewidth]{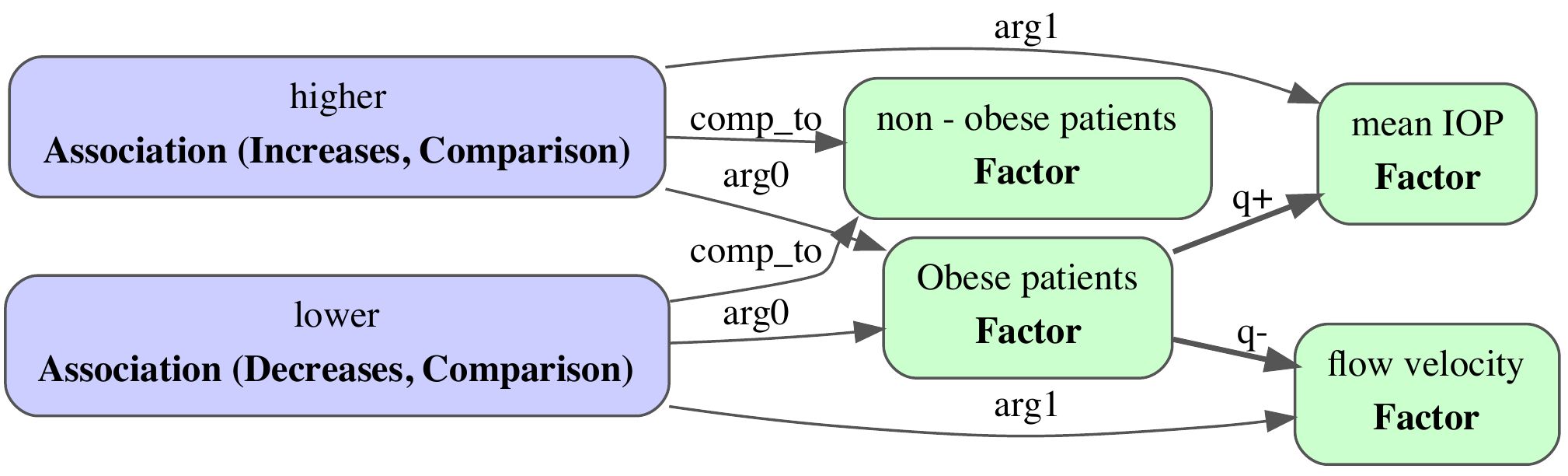}
\caption{\spear knowledge graph output for \textq{Obese patients have a higher mean IOP and lower fow velocity than non-obese patients.} The two qualitative comparisons \textq{higher} and \textq{lower}, support qualitative increase and decrease relations, respectively.}
\label{fig:score3}
\end{figure}

\paragraph{Entities.}
Entities are labeled spans within a textual example.
These are the nodes in the knowledge graph.
The same exact span cannot correspond to more than one entity type, but two entity spans can overlap.
Entities comprise the nodes of Figures \ref{fig:score1}-\ref{fig:score3} upon which attributes and relations are asserted.
Unlike most ontologically-grounded symbolic parsers (e.g., \cite{das2010probabilistic,allen2015complex}), these entity nodes are not ontologically grounded in a class hierarchy; rather, these entity nodes are associated with a token sequence (e.g., \textq{smoking rate} in \figref{score1}) and a corresponding entity class (e.g., \code{Factor}).
These entities also have high-dimensional vectors from the transformer model, which approximates the distributed semantics.

\paragraph{Attributes.}
Attributes are Boolean labels, and each entity (i.e., graph node) may have zero or more associated attributes. 
Attribute inference is therefore a multi-label classification problem.
The previous \spert transformer model was not capable of expressing these; this is a novel contribution of \spear, as described in \secref{arch}.
In Figures \ref{fig:score1}-\ref{fig:score3}, attributes are rendered as parenthetical labels inside the nodes, e.g., \code{Correlation} and \code{Increases} in the \figref{score1} nodes for \textq{associated with} and \textq{higher,} respectively.
The multi-label nature allows the \figref{score1} \textq{higher} node to be categorized simultaneously as \code{Increases} and \code{Comparison}.

\paragraph{Relations.}
Relations are directed edges between labeled entities, representing semantic relationships.
These are critical for expressing what-goes-with-what over the set of entities.
For example in the sentence in \figref{score1}, the relations (i.e., edges) indicate that the \textq{higher} association asserts the antecedent (\code{arg0}) \textq{men} against (\code{comp\_to}) \textq{women} for the consequent (\code{arg1}) \textq{smoking rate.}
Without these relations, the semantic structure of this scientific hypothesis would be ambiguous.
Note that in Figures \ref{fig:score1}-\ref{fig:score3} the unlabeled arrows are all \code{modifier} relations, left blank to avoid clutter.

\subsection{Problem Definition}

We define the multi-attribute knowledge graph extraction task as follows: for a text passage $\mathcal{S}$ of $n$ tokens $s_1,...,s_n$, and a graph schema of entity types $\mathcal{T}_e$, attribute types $\mathcal{T}_a$, and relation types $\mathcal{T}_r$, predict:
\begin{enumerate}
    \item The set of entities $\langle s_j, s_k, t \in \mathcal{T}_e \rangle \in \mathcal{E}$ ranging from tokens $s_j$ to $s_k$, where $0 \leq j \leq k \leq n$,
    \item The set of relations over entities $\langle e_{head} \in \mathcal{E}, e_{tail} \in \mathcal{E}, t \in \mathcal{T}_r \rangle \in \mathcal{R}$ where $e_{head} \neq e_{tail}$,
    \item The set of attributes over entities $\langle e \in \mathcal{E}, t \in \mathcal{T}_a \rangle \in \mathcal{A}$.
\end{enumerate}
This defines a directed multi-graph without self-cycles, where each node has zero to $|\mathcal{T}_a|$ attributes. 
\spear does not presently populate attributes on relations.

\subsection{Knowledge Graph Schemata}

We briefly describe a subset of the graph schemata for our two use-cases: scientific claims and ethnographic mental models.
These two schemata share some qualitative causal representations but vary in other domain-specific descriptions. 
In follow-on work, these schemata may be integrated into a single schema.

\paragraph{Scientific Claims.}
Our scientific claim schema is designed to capture associations between factors (e.g., causation, comparison, prediction, proportionality), monotonicity constraints across factors, epistemic status, and high-level qualifiers.
This model is used for qualitative reasoning to help characterize the replicability and reproducibility of scientific claims \cite{alipourfard2021systematizing,gelman2021toward}.
We describe the entities, attributes, and relations of the schema, referencing the graphed examples rendered by our system in Figures \ref{fig:score1}, \ref{fig:score2}, and \ref{fig:score3}.

This schema includes six entity types:
\textbf{Factors} are variables that are tested or asserted within a claim (e.g., \textq{smoking rate} in \figref{score1});
\textbf{Associations} are explicit phrases associating one or more factors in a causal, comparative, predictive, or proportional assertion (e.g., \textq{associated with} and \textq{reduced} in Figures \ref{fig:score1} and \ref{fig:score2}, respectively);
\textbf{Magnitudes} are modifiers of an association indicating its likelihood, strength, or direction (e.g., \textq{might} and \textq{much} in \figref{score1});
\textbf{Evidence} is an explicit mention of a study, theory, or methodology supporting an association;
\textbf{Epistemics} express the belief status of an association, often indicating whether something is hypothesized, assumed, or observed;
\textbf{Qualifiers} constrain the applicability or scope of an assertion (e.g., \textq{in China} in \figref{score1} and \textq{from 5 February onwards} in \figref{score2}).

This schema includes the following attributes, all of which apply solely to the \textit{Association} entities:
\textbf{Causation} expresses cause-and-effect over its constituent factors (e.g., \textq{reduced} span in \figref{score2});
\textbf{Comparison} expresses an association with a frame of reference, as in the \textq{higher} statement of \figref{score1} and the \textq{higher} and \textq{lower} statements of \figref{score3};
\textbf{Increases} expresses high or increased factor value;
\textbf{Decreases} expresses low or decreased factor value;
\textbf{Indicates} expresses a predictive relationship; and \textbf{Test} indicates a statistical test employed to test a hypothesis.

We encode six relations:
\textbf{arg0} relates an association to its cause, antecedent, subject, or independent variable;
\textbf{arg1} relates an association to its result or dependent variable;
\textbf{comp\_to} is a frame of reference in a comparative association;
\textbf{modifier} relates entities to descriptive elements, e.g., all leftward arrows in Figures \ref{fig:score1} through \ref{fig:score3} (unlabeled for simplicity);
\textbf{q+} and \textbf{q-} indicate positive and negative qualitative proportionality, respectively, where increasing the head factor increases or decreases (the amount or likelihood of) the tail factor, respectively.

\paragraph{Ethnographic Mental Models.}

In our preliminary ethnographic mental modeling domain, we utilize a slightly different schema to capture intentional and functional causality in addition to culturally-specific attributes such as gender and spirituality.

This schema includes attributes for \code{Spirituality} (e.g., \textq{God} and \textq{prayed} in \figref{hab2}), \code{Action/event} (e.g., \textq{prayed} and \textq{free} in \figref{hab2}), \code{Influence} for causally-potent elements (e.g., \textq{prevent} in \figref{hab1}), and others.

We include additional relations \textq{agent/poss} to describe the actor or possessor of an element, temporal precedence \code{t+} relations to indicate one event preceding another, and intentional \code{forPurpose} and functional \code{hasFunction} relations to indicate the goal (i.e., intention or function, respectively) of an action or artifact. 

These relatively simple statements in \figref{hab1} and \figref{hab2} originate from an ethnographic article \cite{aziato2016religious} that includes interview snippets.
Despite their simplicity, the \spear knowledge graphs illustrate rich multi-step causality:
\figref{hab1} indicates that prayer has the purpose of reducing the incidence (or severity of) complications, and \figref{hab2} plots a similar structure for praying for the purpose of preventing pain of the speaker.

\subsection{Scientific Claims Dataset}

Our preliminary dataset for the scientific claims domain consists of 515 examples from Social and Behavior Science (SBS) literature and abstracts from PubMed and the CORD-19 dataset \cite{wang2020cord19}.
Each example consists of a single sentence labeled by a trained NLP expert with one or more spans (possibly nested) identified as entities, zero or more attributes on each entity, and zero or more relations over entities pairs (label counts are listed in \tabref{schema-results}  \textit{support}).
Most datasets for transformer-based information extraction are an order of magnitude larger—and a larger version of our dataset will be validated in the near future—but despite the sparse dataset, our model achieves favorable performance.

\begin{figure}[tb]
\centering
\includegraphics[width=\linewidth]{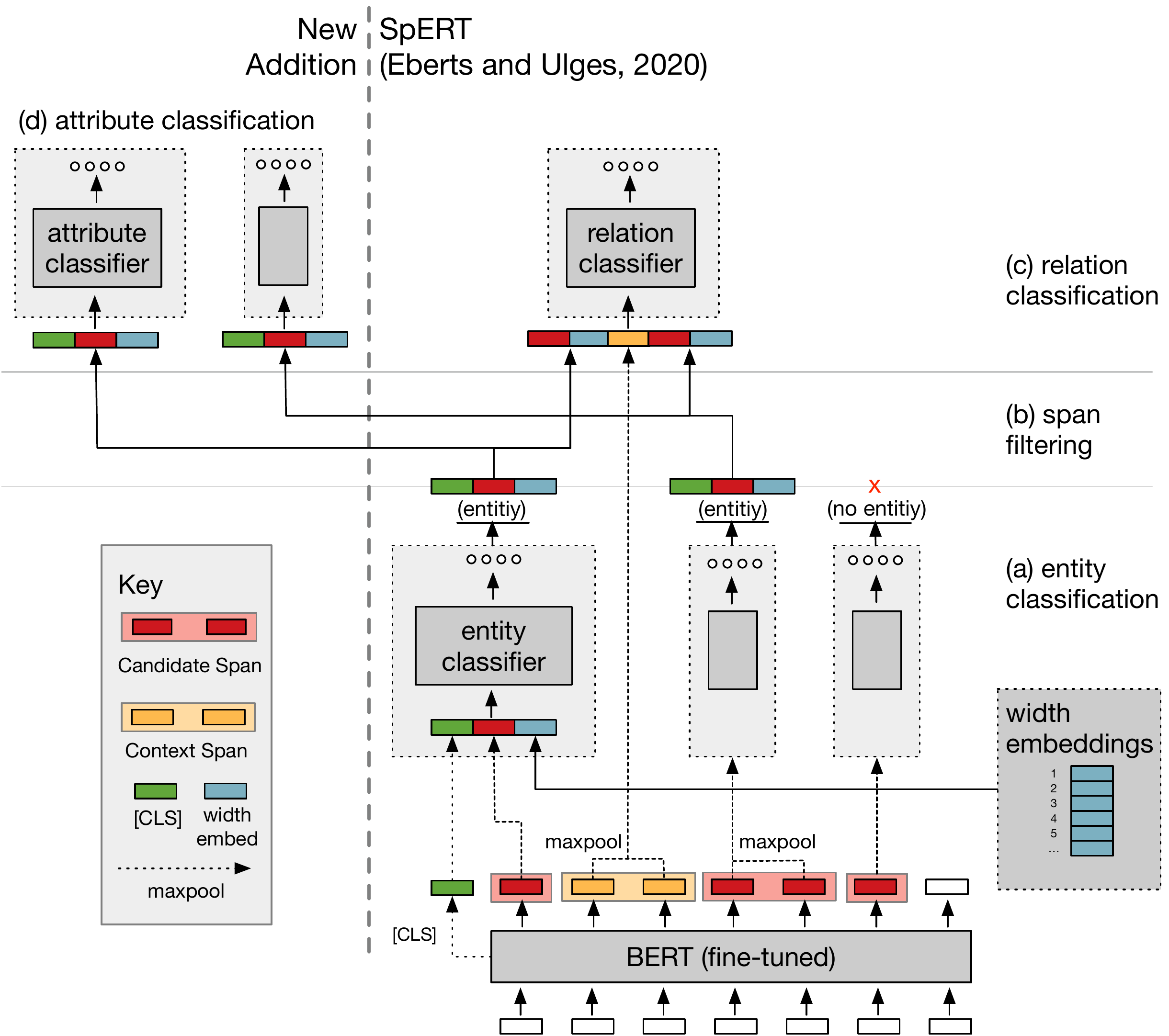}
\caption{Our transformer-based model extends the \spert components (a, b, and c) with attribute classification (d) that performs multi-label inference on identified entity spans.}
\label{fig:model}
\end{figure}

\subsection{Model Architecture}
\label{sec:arch}

Our \spear model architecture extends \spert with an attribute classifier. 
The original architecture provides components (\figref{model} a--c) for joint entity and relation extraction on potentially-overlapping text spans. The parameters of the entity, attribute, and relation classifiers, as well as the parameters of the BERT language model (initialized with its pre-trained values) are all trained end-to-end on our dataset.

The tokens $s_1,...,s_n$ of the text passage $\mathcal{S}$ are each embedded by BERT \cite{devlin-etal-2019-bert} as a sequence $\textbf{e}_1,...,\textbf{e}_n$ of high-dimensional vectors representing the token and its context. BERT also provides an additional \textq{[CLS]} vector output, $\textbf{e}_0$, designed to represent information from the complete text input. For all possible spans,  $span_{j,k} = s_j,..., s_k$, up to a given length, the word vectors associated with a span, $\textbf{e}_j,..., \textbf{e}_k$, are combined by maxpooling to produce a single vector, $\textbf{e}(span_{j,k})$, where each element contains the maximum value across the token vectors for that dimension. The final span representation, $\textbf{x}(span_{j,k})$ is made by concatenating together $\textbf{e}(span_{j,k})$ and $\textbf{e}_0$ along with a width embedding, $\textbf{w}_l$, that encodes the number of words, $l$, in $span_{j,k}$. Each valid span length $l$ looks up a different vector of learned parameters, $\textbf{w}_l$.

The span representation $\textbf{x}(span_{j,k})$ is classified into mutually-exclusive entity types by a linear classifier (\figref{model}a). Only spans identified as entities move on to further analysis (\figref{model}b). All pairings of the remaining entities are classified for relations by a multi-label linear classifier (\figref{model}c), where pairs are represented by the concatenated vectors of the two spans with the \textq{[CLS]} context vector replaced by the maxpool of the token vectors between the entities. 

We implemented the component in (\figref{model}d) to infer multi-label attributes on the identified entities using $\textbf{x}(span_{j,k})$ as input to another multi-label linear classifier. We take only identified entity spans as input to the attribute classifier, as this approach provided best performance and aligns with the finding by \citeauthor{eberts2019span} (\citeyear{eberts2019span}) that training on downstream tasks is best done on strong negative samples consisting of ground truth entities (i.e., teacher forcing).

\begin{figure*}[tb]
\centering
\includegraphics[width=\linewidth]{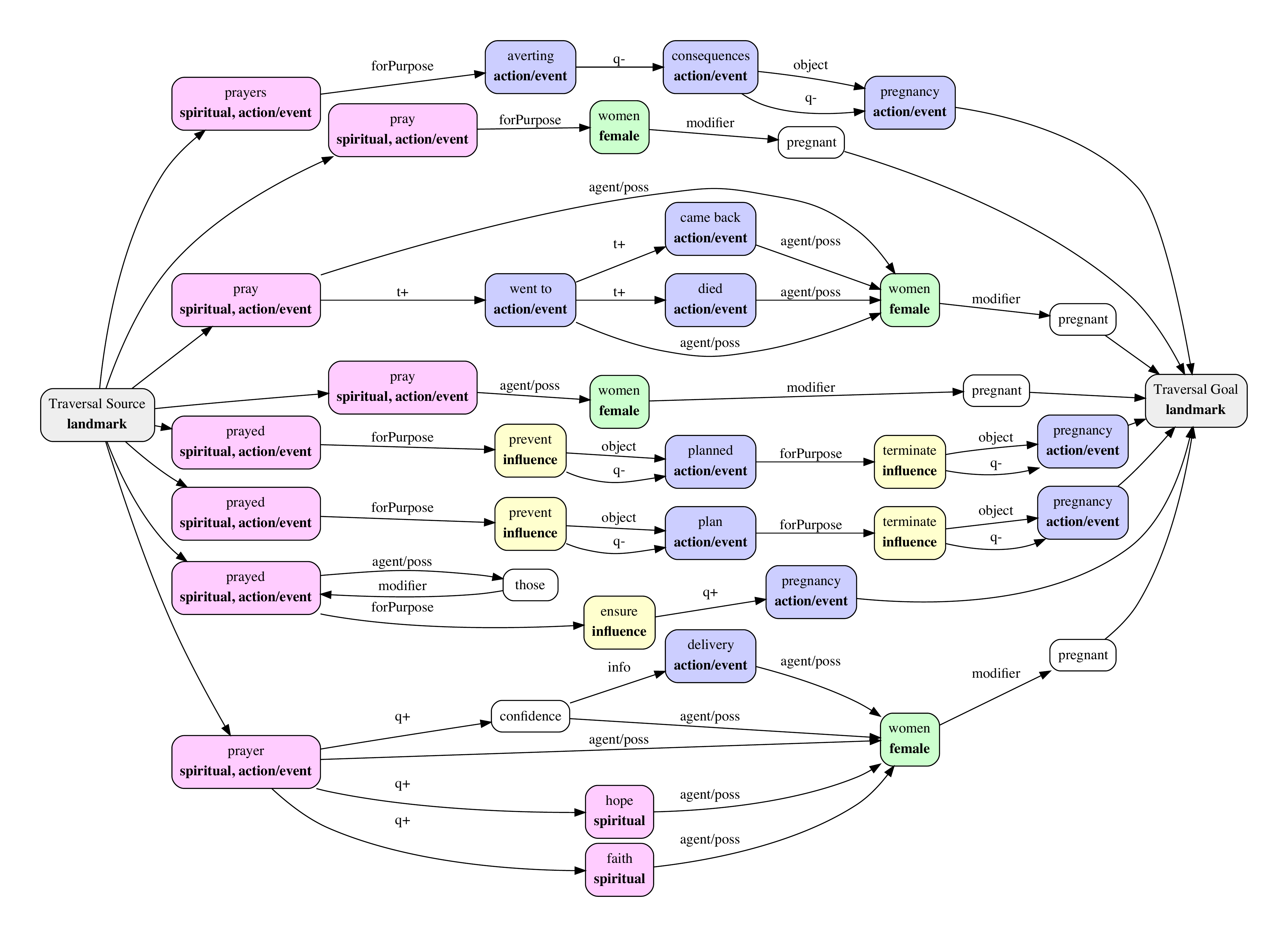}
\caption{A graph traversal from the concept \textq{pray} to the concept \textq{pregnant} after parsing an ethnography about spirituality in pregnancy in Ghana.}
\label{fig:traversal}
\end{figure*}

\begin{table}[htb]
\footnotesize
\centering
\begin{tabular}{c r cccc }
\toprule
& \textbf{Dimension} & \textbf{P} & \textbf{R} & \textbf{F1} & \textbf{Support} \\ \midrule
\parbox{2mm}{\multirow{6}{*}{\rotatebox[origin=c]{90}{\textbf{Entities}}}} 
& factor & 90.13    &    86.71     &   88.39 & 1,604 \\
& evidence & 72.73   &     80.00    &    76.19 & 139 \\
& epistemic & 93.33   &    100.00    &    96.55 & 178\\
& association &  95.89   &     93.33    &    94.59 & 837\\
& magnitude & 94.44    &    94.44     &   94.44 & 415\\
& qualifier & 86.96    &    68.97     &   76.92 & 216 \\
& \textbf{Micro-Averaged} & 91.29    &    87.89    &    89.56 & \\ \midrule
\parbox{2mm}{\multirow{6}{*}{\rotatebox[origin=c]{90}{\textbf{Attributes}}}} 
& causation & 88.24  &      93.75    &    90.91 & 204 \\
& comparison & 79.17  &      90.48   &     84.44 & 234 \\
& indicates & 80.00   &     66.67    &    72.73 & 44 \\
& increases & 75.86    &    95.65    &    84.62 & 262 \\
& decreases & 100.00    &   100.00   &    100.00 & 134 \\
& correlation & 94.74    &    94.74    &    94.74 & 199  \\
& test & 100.00    &    66.67    &    80.00 & 24  \\
& \textbf{Micro-Averaged} & 84.62    &    91.67    &    88.00 & \\\midrule
\parbox{2mm}{\multirow{7}{*}{\rotatebox[origin=c]{90}{\textbf{Relations}}}} 
& arg0 & 82.93   &     76.40    &    79.53 & 865 \\
& arg1 & 76.71    &    71.79    &    74.17 & 883 \\
& comp\_to  & 81.82   &     69.23   &     75.00 & 137 \\
& modifier  & 84.78   &     74.29   &     79.19 & 1,080 \\
& q+ & 77.78    &    56.00    &    65.12 & 295 \\
& q- & 60.00    &    85.71    &    70.59 & 138 \\
& subtype & 85.71    &    75.00    &    80.00 & 106 \\
& \textbf{Micro-Averaged} & 81.00    &    72.97   &     76.78 & \\\bottomrule
\end{tabular}
\caption{Precision, recall, F1 and support (i.e., occurrences in dataset) for each label on 10\% held-out dataset using SpEAR with rectifier and filtering model.}
\label{tab:schema-results}
\end{table}

\section{Results}
\label{sec:results}

We describe two different results of using \spear with our qualitative causal schemata: (1) precision, recall, and F1 measure in the scientific claims domain and (2) traversal through an ethnographic qualitative causal model.
This provides empirical evidence of the effectiveness of our approach and the expressiveness of the qualitative causal schema, respectively.

\subsection{Information Extraction for Scientific Claims}

For our scientific claims dataset, we use the fine-tuned SciBERT transformer variant \cite{beltagy2019scibert} as the input layer of our architecture.

We partitioned our dataset into a randomized 90\% train/test split of 464 and 51 examples, respectively.
We trained our \spear model for 20 epochs and then run our evaluation.
The per-class evaluations are listed in \tabref{schema-results}, divided across the various entities, attributes, and relations. 
\tabref{schema-results} reports the micro-averaged results for entities, attributes, and relationships, as well as support scores to show how many examples of each element are in the full 515-example dataset.
Despite the small size of our preliminary dataset, the model achieves promising results on most classes.

Importantly, the relations and attributes cannot be correct if the entities they are defined over are incorrect.
This means that we expect relations and attributes to have lower precision and recall, all else being equal.
This is especially the case for relations, which require both of their constituent entities (i.e., head and tail nodes) to be properly characterized in order to be scored as correct.
The relations \code{q+} and \code{q-} achieved the lowest performance, due in part to the lower support in the training data, and also due to the greater distance between these spans in the text, all else being equal.

These results support our claim that qualitative causal structure can be characterized by context-sensitive NLP models.

\subsection{Traversing Ethnographic Causal Models}

In the ethnographic domain, we trained \spear on labeled examples from Anthropology papers describing religious beliefs surrounding pregnancy in Ghana \cite{aziato2016religious}.
We then ran \spear to extract information from these and other sentences from the same literature, resulting in a global causal graph comprising the disconnected knowledge graphs from each sentence from the literature.
These preliminary results include the use of human-labeled training data, so we consider this a proof-of-concept study of the practicality of the causal structure.

We then built a traversal system to walk to and/or from any concept in this global causal graph, along the nodes and edges extracted by \spear.
The source and destination concepts are given by the user, and then the system identifies all source and destination nodes by vector- or lemma-distance to the user's inputs.
Given these source and destination nodes, it identifies global paths from one concept to the other.

The traversal produces a graph such as the one in \figref{traversal}, which shows paths from the source \textq{pray} (including \textq{prayed,} \textq{prayer,} \textq{prayers,} etc.) to the destination \textq{pregnant} (also including \textq{pregnancy.})
These traversals describe prayer for the purpose of preventing unfortunate consequences, ensuring safe pregnancy, and qualitatively increasing faith, hope, and confidence in delivery. 
These graph structures only contain complete paths, so the extraneous structure has been pruned for readability at the expense of completeness.

These results support our claim that the causal models extracted by our transformer-based NLP can support coarse-level reasoning and traversal across concepts.

\section{Conclusion}
\label{sec:conclusion}

This paper describes a transformer-based NLP model for extracting entities, attributes, and relationships that describe qualitative causal structure.
We demonstrated the approach in the two different domains of scientific claims and descriptions of mental models from ethnography.
Our datasets are still under development, but despite their relative sparsity they support encouraging results with respect to F1-measure and traversal.

One limitation of this work is that the nodes are not grounded in a formal hierarchical ontology.
This means that many of the assumptions about the arguments to \code{q+} and \code{q-} may not hold in \spear's output: \code{q+} may be expressed over quantities, over events, over adjectives, or any heterogeneous mix of these, and a downstream reasoner has no formal \emph{a priori} indicator of which these are.
One remedy to this is to use our node-based attributes to express these different types of elements, but whether the transformer-based NLP model can accurately classify these abstract categories is an empirical question.

Our near-term future work is to expand our datasets and utilizing \spear's results in downstream systems, e.g., for estimating the reproducibility of a scientific claim, automatically organizing and combining insights from academic literature, and globally traversing descriptive mental models to identify culturally-specific causally-potent concepts and purposes.

\hide{
\section*{Acknowledgments}

The preparation of these instructions and the \LaTeX{} and Bib\TeX{}
files that implement them was supported by Schlumberger Palo Alto
Research, AT\&T Bell Laboratories, and Morgan Kaufmann Publishers.
Preparation of the Microsoft Word file was supported by IJCAI.  An
early version of this document was created by Shirley Jowell and Peter
F. Patel-Schneider.  It was subsequently modified by Jennifer
Ballentine and Thomas Dean, Bernhard Nebel, Daniel Pagenstecher,
Kurt Steinkraus, Toby Walsh and Carles Sierra. The current version
has been prepared by Marc Pujol-Gonzalez and Francisco Cruz-Mencia.

}

\bibliographystyle{named}
\bibliography{ijcai21,main}

\end{document}